\PassOptionsToPackage{table}{xcolor}
\documentclass{article}
\usepackage{iclr2027_conference,times}
\iclrfinalcopy 

\usepackage{amsmath,amsfonts,bm}

\def\eqref#1{equation~\ref{#1}}

\def\1{\bm{1}}

\DeclareMathAlphabet{\mathsfit}{\encodingdefault}{\sfdefault}{m}{sl}
\SetMathAlphabet{\mathsfit}{bold}{\encodingdefault}{\sfdefault}{bx}{n}

\usepackage{hyperref}
\usepackage{url}
\usepackage{graphicx}
\usepackage{booktabs}
\usepackage{amsmath}
\usepackage{amssymb}
\usepackage{multirow}
\usepackage{xcolor}
\usepackage{colortbl}
\usepackage{float}
\usepackage{placeins}
\usepackage{enumitem}
\usepackage{array}
\graphicspath{{figures/}}
\definecolor{oursA}{RGB}{226,239,247}
\definecolor{oursB}{RGB}{253,231,229}
\definecolor{evid}{RGB}{0,128,0}
\definecolor{miss}{RGB}{200,0,0}
\newcommand{\method}{TriageRAG}
\newcommand{\pcb}{p_{\mathrm{cb}}}
\newcommand{\pob}{p_{\mathrm{ob}}}
\newcommand{\cat}[1]{\textsc{#1}}
\newcommand{\evid}[1]{\textcolor{evid}{#1}}
\newcommand{\miss}[1]{\textcolor{miss}{#1}}

\newcommand{\ci}[2]{{\scriptsize[#1, #2]}}

\title{TriageRAG: Failure-Aware Self-Training for Adapting Retrieval-Augmented Language Models to Specialized Domains}

\author{Samarth Varshney \And Ritu Yadav}

\begin{document}

\maketitle
\lhead{Preprint}

\begin{abstract}
Retrieval-augmented generation (RAG) lets large language models (LLMs) answer knowledge-intensive questions from retrieved evidence, but adapting a general-purpose RAG system to a specialized domain such as medicine is difficult when only an unlabeled corpus is available. Self-training approaches such as SimRAG address this by letting the LLM write its own question--answer pairs from domain passages and fine-tuning on the pairs that survive a round-trip consistency filter. We observe that these approaches treat every retained example identically: they never ask whether the student already knows the answer, already extracts it correctly from the retrieved context, or fails---and if so, whether the failure lies with the reader or with the retriever. To tackle this, we propose \method{}, a failure-aware self-training approach that probes the student on each self-generated question with and without retrieved context, triages it into one of four categories (\cat{known}, \cat{solved}, \cat{context-failure}, \cat{retrieval-failure}), and routes each category to the training signal that addresses it, including a parametric route that teaches the reader to answer when retrieval misses. Probing is cheap and repeated every round, so the training mixture follows the student, and the triage histogram doubles as a label-free diagnostic of where the domain gap lies. Experiments on 11 datasets spanning three domains with a Qwen3-4B backbone, and on CS-Bench with Qwen3-8B, demonstrate that \method{} outperforms a budget-matched SimRAG by 1.2--6.4 points and is the only method that recovers the closed-book level in every case where retrieval hurts the base model. A single ablation identifies the mechanism: round-trip filtering removes every training context that lacks the evidence, and a reader that never sees such contexts learns to over-trust noisy retrieval.
\end{abstract}
\section{Introduction}

Retrieval-augmented generation (RAG) \cite{lewis2020rag,gao2023survey,asai2024selfrag} grounds the answers of large language models (LLMs) in retrieved evidence, which adapts them to long-tail knowledge without retraining \cite{ovadia2023finetuning} and reduces hallucination \cite{shuster2021retrieval}. Adapting a general-purpose RAG system to a specialized domain such as medicine is nevertheless hard: reader and retriever both face a distribution shift, and labeled domain QA data is scarce or private, so fine-tuning has to rely on synthetic data. SimRAG \cite{xu2025simrag} is a strong recent recipe. It fine-tunes an instruction-tuned LLM on general retrieval-oriented data, prompts the same model to write short-span, multiple-choice and claim-verification questions from unlabeled domain passages, keeps the pairs whose source passage is retrieved in the top-$k$ for the generated question (round-trip consistency), and fine-tunes on the survivors with retrieved contexts.

We start from an observation about what such a pipeline does \emph{not} look at: every retained pseudo-labeled pair receives the same treatment. Nothing in the loop asks whether the student already knows the answer without context, whether it already extracts it correctly from the retrieved passages, or whether it fails---and if so, whether the failure belongs to the reader (the evidence is in the context but the answer is wrong) or to the retriever (the evidence never reached the context). These problems have different remedies: contextual fine-tuning teaches a reader to use evidence that is present, but it cannot help when the evidence is absent. Worse, the round-trip filter removes exactly the pairs whose evidence is absent, so the fine-tuned reader never sees a context that lacks the answer and, in a domain where the retriever misses often, learns to trust whatever it is given.

\begin{figure*}[t]
\centering
\includegraphics[width=0.74\textwidth]{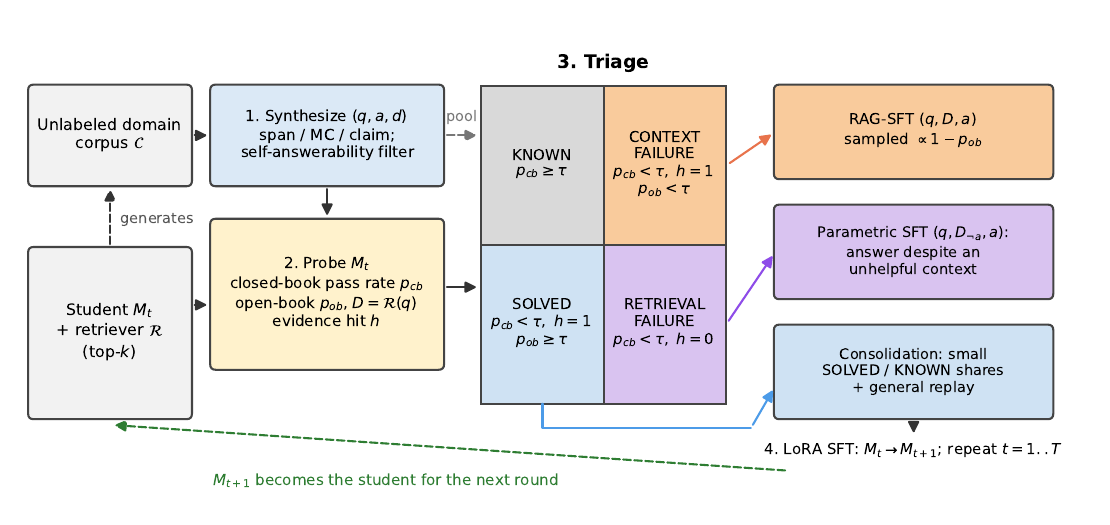}
\caption{Overview of \method{}. Each round, the student synthesizes $(q,a,d)$ triples from the unlabeled corpus (span, multiple-choice and claim formats; only self-answerable pairs are kept), is probed closed-book ($\pcb$) and open-book on the top-$k$ retrieved passages ($\pob$, evidence hit $h$), and each pair is triaged by the conditions shown and routed to a training signal; the routed mixture is used for LoRA fine-tuning and the updated student re-probes the pool in the next round. The retriever is frozen in our main configuration (Section~\ref{sec:retriever}).}
\label{fig:framework}
\end{figure*}

We propose \method{}\footnote{Failure-aware, \emph{triage}-based self-training for RAG.}, which lets the student's own behaviour decide how each self-generated example is used (Figure~\ref{fig:framework}). After synthesizing question--answer pairs as in SimRAG, \method{} \emph{probes} the student on every pair with and without retrieved context, \emph{triages} it into \cat{known}, \cat{solved}, \cat{context-failure} or \cat{retrieval-failure}, and \emph{routes} each category to a different treatment: contextual supervised fine-tuning (SFT) for context failures; a \emph{parametric} route that trains the reader to produce the verified answer despite an unhelpful retrieved context---the situation it faces whenever retrieval misses---for retrieval failures; and small consolidation shares for the rest. Probing costs a fraction of fine-tuning, so the diagnosis is repeated every round and the mixture follows the student. The triage histogram is itself a label-free diagnostic of where the domain gap lies.

We evaluate on 11 datasets from medicine, natural science and computer science (CS) with a Qwen3-4B backbone, and on CS with Qwen3-8B as well, against SimRAG, a RAFT-style mixture and a random-selection control that share the same Stage-I model, synthetic pool, filter, recipe and budget. Our contributions are:
\begin{itemize}[leftmargin=*,itemsep=1pt,topsep=2pt]
\item A student-aware view of self-training for domain RAG in which each self-generated question is classified by \emph{failure type} with cheap closed-book and open-book probes, yielding a label-free diagnostic of the domain gap.
\item \method{}, which routes the four categories to different training signals and iterates, with SFT only and no reward model, judge or labeled data.
\item Experiments on three domains and two backbone sizes in which \method{} outperforms a budget-matched SimRAG by 1.2--6.4 points, with three training seeds and paired bootstrap intervals on every headline comparison; an ablation that identifies the parametric route as the ingredient behind the gap; and a controlled negative result on tuning the retriever with retrieval failures.
\end{itemize}

\section{Related Work}

\paragraph{Retrieval-augmented generation.}
RAG is a standard tool for knowledge-intensive tasks \cite{lewis2020rag,gao2023survey}; recent work refines when to retrieve \cite{jiang2023flare,asai2024selfrag}, makes readers robust to irrelevant passages \cite{yoran2024robust,shi2023distracted,cuconasu2024noise} and instruction-tunes LLMs for RAG \cite{lin2024radit,liu2024chatqa}. For specialized domains, RAFT \cite{zhang2024raft} fine-tunes a reader on GPT-4-written answers with the oracle document and distractors, including distractor-only examples, and \citet{schimanski2024evidence} distill context-aware QA data from GPT-4. SimRAG \cite{xu2025simrag} removes the proprietary teacher by giving the student question-answering and question-generation skills and keeping round-trip-consistent pairs \cite{bartolo2021roundtrip}. \method{} keeps SimRAG's self-generation and adds the missing piece: the student's behaviour decides how each generated example is used.

\paragraph{Self-training and self-improvement.}
Self-training fits a student on teacher-produced labels and has been applied to LLM instruction tuning \cite{wang2023selfinstruct,li2024backtranslation}, alignment \cite{yuan2024selfrewarding,gulcehre2023rest} and, in SimRAG, to RAG; its main weakness is label noise. Recent self-play methods co-evolve a challenger and a solver and use the solver's pass rate to keep tasks near its frontier \cite{huang2025rzero,kuba2025lsp,lu2025ssp,kwan2026scope}; they optimize task \emph{difficulty} for reinforcement learning of reasoning agents, whereas \method{} classifies the \emph{type} of failure of a corpus-grounded RAG system to choose among training signals, with plain SFT.

\paragraph{Knowledge-aware fine-tuning.}
\citet{gekhman2024unknown} and \citet{kang2024unfamiliar} show that fine-tuning on facts the model does not know encourages hallucination; KaFT \cite{chen2025kaft} weights domain SFT examples by the conflict between parametric and contextual knowledge; \citet{longpre2021conflicts} and \citet{mallen2023trust} study when to trust memory over context. These works use labeled data. \method{} applies knowledge probing to \emph{self-generated} data and adds the dimension that matters for RAG---whether the evidence was retrieved---so that reader and retriever failures are treated differently.

\section{Methodology}
\label{sec:method}

\subsection{Problem Setup}
In a RAG problem, a retriever $\mathcal{R}$ returns the top-$k$ passages $D=\{d_1,\dots,d_k\}$ from a corpus $\mathcal{C}$ for a query $q$, and the LLM $\mathcal{M}_\theta$ generates an answer $a$ from $(q, D)$. We aim to improve the QA capability of $\mathcal{M}_\theta$ in a specialized domain where only the unlabeled corpus $\mathcal{C}$ is available. Following SimRAG, we first fine-tune the model on retrieval-oriented general-domain data (Stage-I) and then adapt it to the domain with self-generated pseudo-labeled tuples (Stage-II). \method{} changes Stage-II: instead of treating all retained tuples identically, it diagnoses the student on each tuple and routes the tuple to the training signal that fits the diagnosis. The retriever is kept frozen in the main configuration; Section~\ref{sec:retriever} discusses a retriever route.

\subsection{Stage-I: Retrieval-oriented Fine-tuning}
We fine-tune an instruction-tuned backbone on a scaled-down version of SimRAG's Stage-I blend (16,000 examples; Appendix~\ref{app:stage1}): general context-aware QA in short-span, yes/no and multiple-choice formats, and the two skills used in synthesis, \emph{answer generation} (candidate answer spans from a passage) and \emph{answer-conditioned question generation}.

\subsection{Stage-II: Failure-Aware Self-Training}
Each round $t$ takes the current student $\mathcal{M}_t$ ($\mathcal{M}_1$ is the Stage-I model) through four steps.

\paragraph{(1) Synthesis.} For a sampled passage $d\in\mathcal{C}$ the student proposes candidate answer spans $a\sim\mathcal{M}_t(\cdot\mid d)$ and then an answer-conditioned question $q\sim\mathcal{M}_t(\cdot\mid a,d)$. As in SimRAG we generate three formats: short-span questions; multiple-choice questions, which reuse a span question and add three distractor spans drawn from other passages, matched on whether the answer is numeric; and claims that are supported or refuted by $d$. We keep a tuple $(q,a,d)$ only if the student recovers $a$ when given the gold passage $d$ (\emph{self-answerability}). This label-noise filter is applied identically to every method we compare. We do \emph{not} gate on round-trip retrieval consistency: whether the source passage is retrieved is recorded as a signal and used by the triage. The synthesis is run once per domain and the pool is shared by all methods and rounds.

\paragraph{(2) Probing.} For every retained tuple we draw $n$ samples from the student with no context and $n$ samples with the top-$k$ passages $D=\mathcal{R}(q)$ retrieved with the question alone, as at test time, and compute the closed-book and open-book pass rates $\pcb = \tfrac{1}{n}\sum_{i}\mathbb{1}[\hat a_i^{\mathrm{cb}}\equiv a]$ and $\pob = \tfrac{1}{n}\sum_{i}\mathbb{1}[\hat a_i^{\mathrm{ob}}\equiv a]$, where $\equiv$ is exact match for letters and yes/no answers and a normalized token-overlap match for spans. The evidence hit is $h=\mathbb{1}[d\in D\ \lor\ a\subseteq\mathrm{text}(D)]$.

\paragraph{(3) Triage.} With a threshold $\tau$ ($0.75$ throughout) each tuple falls into exactly one category (the tests are applied in this order):
\begin{itemize}[leftmargin=*,itemsep=0pt,topsep=1pt]
\item \cat{known}: $\pcb\ge\tau$; the student already knows the answer without context.
\item \cat{solved}: $\pcb<\tau$, $h=1$, $\pob\ge\tau$; the reader uses the retrieved evidence correctly.
\item \cat{context-failure}: $\pcb<\tau$, $h=1$, $\pob<\tau$; the evidence is in the context but the reader fails or is unstable.
\item \cat{retrieval-failure}: $\pcb<\tau$, $h=0$; the evidence was not retrieved, so no contextual training can make the question answerable from $D$.
\end{itemize}
The category histogram is a diagnostic of the domain gap (Section~\ref{sec:triage_study}).

\paragraph{(4) Routing and training.} Given a budget of $B$ Stage-II examples per round, we fill per-category quotas of 60\%/20\%/15\%/5\% for \cat{context-failure}/\cat{retrieval-failure}/\cat{solved}/\cat{known}; an unfilled quota is topped up from the remaining context failures, then solved items, without duplicates. In our pools the \cat{context-failure} category is smaller than its quota in every domain, so every context failure is used and the shortfall is filled with \cat{solved} items (realized shares in Appendix~\ref{app:baselines}). Each category receives a different treatment:
\begin{itemize}[leftmargin=*,itemsep=0pt,topsep=1pt]
\item \textbf{Contextual SFT} for \cat{context-failure}: the target is $a$ given $(q,D)$; when the category exceeds its quota, tuples are sampled without replacement with probability $\propto 1-\pob$, so the items the student gets most wrong are preferred.
\item \textbf{Parametric SFT} for \cat{retrieval-failure}: the target is $a$ given $(q,D)$ where $D$ does \emph{not} contain the evidence. The answer is verified against the source passage, so the label is trustworthy, but the reader must produce it from parametric knowledge and learns that a retrieved context can be unhelpful---the condition it meets at test time whenever retrieval misses.
\item \textbf{Consolidation} for \cat{solved} and \cat{known}: contextual SFT at small shares, plus a fixed replay of general-domain context-QA examples, to preserve answer formats and general skills.
\end{itemize}
The mixture is used for one epoch of LoRA fine-tuning \cite{hu2022lora}; the merged model becomes $\mathcal{M}_{t+1}$ and re-probes the pool in the next round.

\subsection{A Retriever Route}
\label{sec:retriever}
\cat{retrieval-failure} tuples are also natural retriever training data: $(q,d)$ is a positive pair and the top-ranked retrieved passage is a mined hard negative. In a variant of \method{} we fine-tune the dense retriever on these pairs with a contrastive loss after each round and re-index. Section~\ref{sec:ablation} shows that this degrades both retrieval and accuracy, so our main configuration keeps the retriever frozen.

\section{Experimental Setup}
\label{sec:setup}

\subsection{Tasks and Datasets}
We evaluate on 11 datasets from three domains (details in Appendix~\ref{app:testdata}). For the \textbf{medical} domain we use the five datasets of the MIRAGE benchmark \cite{xiong2024mirage}---PubMedQA \cite{jin2019pubmedqa}, BioASQ \cite{tsatsaronis2015bioasq}, MedQA \cite{jin2021medqa}, MedMCQA \cite{pal2022medmcqa} and the medical subsets of MMLU \cite{hendrycks2021mmlu}---and add the text subset of MedXpertQA \cite{zuo2025medxpertqa}, a recent expert-level benchmark with up to ten options. For the \textbf{scientific} domain we use ARC-Challenge \cite{clark2018arc}, SciQ \cite{welbl2017sciq} as four-way multiple choice, and the 14 scientific subsets of MMLU used by SimRAG. For \textbf{computer science} we use the English multiple-choice (MC) and assertion (AS) questions of CS-Bench \cite{song2024csbench}, which can be scored automatically. All questions are multiple-choice, yes/no or yes/no/maybe, and accuracy is the metric throughout.

The unlabeled corpora, which also serve as the retrieval index at test time, are MedRAG textbook snippets plus a PubMed slice for medicine (950,746 passages), SciQ supports and OpenBookQA facts for science (10,863) and the Wikipedia computer-science subset used by SimRAG for CS (6,237); see Appendix~\ref{app:testdata}.

\subsection{Baselines}
Because our question is how self-generated data should be \emph{used}, all Stage-II methods share the Stage-I model, the synthesized pool, the self-answerability filter, the budget $B$, the general replay and the training recipe, and differ only in which examples are selected and how their contexts are constructed. (1) \emph{No domain adaptation}: the backbone closed-book, the backbone with RAG, and the Stage-I model with RAG (equivalent to ``SimRAG w/o Stage-II''). (2) \emph{Self-training baselines at equal budget}: SimRAG$^\ddagger$, our re-implementation, which keeps tuples whose source passage is among the top-$k$ (round-trip consistency) and samples them uniformly with retrieved contexts; \emph{Random selection}, which samples $B$ tuples uniformly from the full pool with retrieved contexts and isolates the effect of selection from that of volume; and RAFT-style$^\ddagger$ \cite{zhang2024raft}, which places the gold passage among $k{-}1$ retrieved distractors for 80\% of the examples and uses distractors only for the remaining 20\%. (3) \emph{Ablations} of \method{} (Section~\ref{sec:ablation}). We do not compare with GPT-4-distilled readers or with domain LLMs trained on task data overlapping our test sets; all rows use the same retriever and prompt.

\subsection{Implementation Details}
Backbones are Qwen3-4B-Instruct-2507 and Qwen3-8B in non-thinking mode \cite{qwen3}. All fine-tuning uses LoRA ($r{=}64$, $\alpha{=}128$, all linear layers), one epoch, learning rate $10^{-4}$ with cosine decay, global batch 64 and maximum length 3,072. The retriever is bge-base-en-v1.5 \cite{bge} with a flat FAISS index \cite{johnson2019faiss} and $k{=}5$ for probing, training and evaluation. Synthesis uses 12,000 medical, 8,000 science and all 5,089 CS passages of at least 40 words at temperature 0.7, yielding pools of 40,540, 32,686 and 20,669 self-answerable tuples. Probing uses $n{=}4$ samples per condition at temperature 0.7 and $\tau{=}0.75$. The per-round budget is $B{=}12{,}000$ for medicine and $6{,}000$ for science and CS (30\%, 18\% and 29\% of the respective pools, so that selection matters), plus 2,000 general replay examples. \method{} runs $T{=}3$ rounds; SimRAG runs one round as in the original paper and, as a control, three. Evaluation is zero-shot with greedy decoding; parse failures count as wrong (Appendix~\ref{app:parse}). One medical round of \method{} (4B) takes about 15 minutes of probing with vLLM \cite{kwon2023vllm}, 80 minutes of LoRA training (two A100s) and 10 minutes of evaluation. Prompts are in Appendix~\ref{app:prompts}.

\section{Experimental Results}
\label{sec:results}

\begin{table*}[t]
\centering\small
\setlength{\tabcolsep}{5pt}
\resizebox{\textwidth}{!}{\begin{tabular}{lccccccc}
\toprule
\textbf{Datasets} & \textbf{PubMedQA} & \textbf{BioASQ} & \textbf{MedQA} & \textbf{MedMCQA} & \textbf{MMLU-Med} & \textbf{MedXpertQA} & \textbf{Avg.} \\
\textbf{\# questions} & 500 & 618 & 1,273 & 4,183 & 1,089 & 2,450 & --- \\
\midrule
\multicolumn{8}{l}{\textit{Backbone: Qwen3-4B-Instruct}} \\
Backbone (closed-book) & 31.00 & 77.67 & 60.49 & 54.24 & 76.03 & \underline{14.53} & 52.33 \\
Backbone + RAG & 68.20 & 92.07 & 61.35 & 56.94 & 76.49 & \textbf{14.82} & 61.65 \\
Stage-I + RAG & \underline{82.00} & \underline{92.56} & 61.12 & 59.22 & 75.02 & 13.84 & 63.96 \\
\cmidrule(lr){1-8}
SimRAG$^\ddagger$ & \textbf{82.60} & 91.26 & 59.86 & 59.62 & 74.93 & 13.76 & 63.67 \\
SimRAG$^\ddagger$, 3 rounds & 80.60 & 91.59 & 58.99 & 59.12 & 74.47 & 13.27 & 63.01 \\
Random selection & 77.60 & \textbf{92.88} & 61.51 & \underline{61.92} & 77.13 & 13.84 & 64.15 \\
RAFT-style$^\ddagger$ & 79.00 & 92.39 & 61.04 & 61.44 & 77.04 & 13.67 & 64.10 \\
\cmidrule(lr){1-8}
\rowcolor{oursA} \method{} & 81.60 & 91.26 & \textbf{63.08} & 61.37 & \underline{78.33} & 13.39 & \textbf{64.84} \\
\method{}, 2 rounds & 81.60 & 91.75 & 61.19 & 61.18 & \textbf{78.88} & 13.84 & \underline{64.74} \\
\method{}, 3 rounds & 80.80 & 91.26 & \underline{61.90} & \textbf{62.20} & 77.32 & 13.55 & 64.50 \\
\bottomrule
\end{tabular}
}
\caption{Results in the medical domain (accuracy, \%). All methods use RAG with the same bge-base retriever and top-5 passages at inference. \textbf{Bold} and \underline{underline} mark the best and second-best result per column. $^\ddagger$: our own implementation with the same backbone, pool and budget. All Stage-II methods share the Stage-I model, the 40,540-tuple self-generated pool, a budget of 12,000 examples per round and 2,000 general replay examples. The notations are the same for the following tables.}
\label{tab:medical}
\end{table*}

\begin{table*}[t]
\centering\footnotesize
\begin{minipage}[t]{0.42\textwidth}
\centering
\setlength{\tabcolsep}{4pt}
\resizebox{\linewidth}{!}{\begin{tabular}{lcccc}
\toprule
\textbf{Datasets} & \textbf{ARC-C} & \textbf{SciQ} & \textbf{MMLU-Sci} & \textbf{Avg.} \\
\textbf{\# questions} & 1,172 & 1,000 & 3,013 & --- \\
\midrule
\multicolumn{5}{l}{\textit{Backbone: Qwen3-4B-Instruct}} \\
Backbone (closed-book) & 89.25 & 94.60 & \underline{78.29} & 87.38 \\
Backbone + RAG & 87.37 & 94.40 & 74.18 & 85.32 \\
Stage-I + RAG & 88.99 & 94.50 & 74.58 & 86.02 \\
\cmidrule(lr){1-5}
SimRAG$^\ddagger$ & 88.40 & 94.20 & 74.61 & 85.74 \\
Random selection & \underline{89.76} & 95.70 & 77.07 & 87.51 \\
\cmidrule(lr){1-5}
\rowcolor{oursA} \method{} & 89.51 & 95.60 & 77.76 & \underline{87.62} \\
\method{}, 2 rounds & \textbf{89.85} & \underline{95.80} & \textbf{78.33} & \textbf{87.99} \\
\method{}, 3 rounds & 89.16 & \textbf{96.00} & 77.33 & 87.50 \\
\bottomrule
\end{tabular}
}
\caption{Results in the scientific domain (Qwen3-4B; budget 6,000 per round; 32,686-tuple pool).}
\label{tab:science}
\end{minipage}\hfill
\begin{minipage}[t]{0.56\textwidth}
\centering
\setlength{\tabcolsep}{4pt}
\resizebox{\linewidth}{!}{\begin{tabular}{lccc}
\toprule
\textbf{Datasets} & \textbf{MC} & \textbf{AS} & \textbf{Avg.} \\
\textbf{\# questions} & 1,336 & 442 & --- \\
\midrule
\multicolumn{4}{l}{\textit{Backbone: Qwen3-4B-Instruct}} \\
Backbone (closed-book) & \textbf{68.79} & 67.19 & 67.99 \\
Backbone + RAG & 62.05 & 65.16 & 63.60 \\
Stage-I + RAG & 61.60 & 62.22 & 61.91 \\
\cmidrule(lr){1-4}
SimRAG$^\ddagger$ & 61.23 & 62.44 & 61.83 \\
SimRAG$^\ddagger$, 3 rounds & 60.93 & 64.71 & 62.82 \\
Random selection & 67.29 & 67.87 & 67.58 \\
RAFT-style$^\ddagger$ & 67.37 & 67.19 & 67.28 \\
\cmidrule(lr){1-4}
\rowcolor{oursA} \method{} & 66.92 & \underline{69.46} & 68.19 \\
\method{}, 2 rounds & \underline{67.89} & \textbf{69.68} & \textbf{68.78} \\
\method{}, 3 rounds & 67.14 & \underline{69.46} & \underline{68.30} \\
\midrule
\multicolumn{4}{l}{\textit{Backbone: Qwen3-8B}} \\
Backbone (closed-book) & \textbf{70.51} & \underline{69.00} & \underline{69.75} \\
Backbone + RAG & 63.25 & \textbf{71.49} & 67.37 \\
Stage-I + RAG & 64.30 & 67.42 & 65.86 \\
\cmidrule(lr){1-4}
SimRAG$^\ddagger$ & 63.70 & 67.65 & 65.68 \\
Random selection & 68.56 & 66.97 & 67.77 \\
\cmidrule(lr){1-4}
\rowcolor{oursB} \method{} & \underline{68.64} & \textbf{71.49} & \textbf{70.06} \\
\method{}, 2 rounds & 68.04 & 67.42 & 67.73 \\
\bottomrule
\end{tabular}
}
\caption{Results in the computer-science domain (CS-Bench; MC = multiple choice, AS = assertion; budget 6,000 per round; 20,669-tuple pool) with two backbones.}
\label{tab:cs}
\end{minipage}
\end{table*}

\subsection{Main Results}
Tables~\ref{tab:medical}, \ref{tab:science} and \ref{tab:cs} present the results for the medical, scientific and computer-science domains; we make four observations.

(1) \emph{\method{} consistently outperforms budget-matched self-training baselines.} Against SimRAG at the same budget it gains 1.2 points on average in medicine, 1.9 in science and 6.4 on CS-Bench with the 4B backbone, and 4.4 on CS-Bench with the 8B backbone; it is the best fine-tuned method on average in every domain and the best or second best on 8 of the 11 datasets.

(2) \emph{Retrieval is not uniformly helpful, and self-training on retrieved contexts can lock in its harm.} In medicine, five retrieved passages add 9.3 points to the backbone, almost entirely on the literature-grounded PubMedQA and BioASQ. In science and CS the backbone \emph{loses} 2.1 and 4.4 points when given five passages (87.4$\rightarrow$85.3, 68.0$\rightarrow$63.6), and neither Stage-I nor SimRAG repairs the loss; \method{} is the only method that returns to or exceeds the closed-book level in a single round in all three cases (87.6 vs.\ 87.4; 68.2 vs.\ 68.0; 70.1 vs.\ 69.8 for Qwen3-8B), random selection does so in science only (87.5). In medicine its gain over Stage-I (+0.9) is concentrated on the exam-style datasets (MedQA +2.0, MedMCQA +2.2, MMLU-Med +3.3), while the literature yes/no datasets, on which retrieval already supplies the answer, move by at most 1.3 points. MedXpertQA stays near the ten-way chance level for every configuration and does not separate methods at this scale.

(3) \emph{Selection matters less than what the selected contexts look like.} The random-selection control differs from SimRAG only in dropping the round-trip filter, yet it is 5.7, 1.8 and 0.5 points above SimRAG on CS, science and medicine (paired bootstrap over questions: $+5.7$ \ci{3.4}{8.1} on CS, $+1.8$ \ci{1.2}{2.4} on science) and within 0.1--0.7 of \method{} with the 4B backbone. The triage's own contribution over random selection is therefore small at 4B---$+0.6$ \ci{-0.9}{2.1} on CS, $+0.1$ \ci{-0.4}{0.6} on science, $+0.7$ \ci{0.1}{1.3} in medicine---and becomes clear at 8B, where \method{} leads random selection by $+2.3$ \ci{0.2}{4.4} on CS-Bench and on every training seed (Section~\ref{sec:seeds}). Two effects are at work---whether the mixture contains evidence-absent contexts (random selection, RAFT-style and \method{} do; SimRAG does not) and which questions are chosen---which Section~\ref{sec:ablation} isolates.

(4) \emph{Iterating helps \method{} more than SimRAG, but a fixed pool saturates quickly.} \method{} improves in round two on science and CS (87.6$\rightarrow$88.0, 68.2$\rightarrow$68.8) and plateaus in medicine (64.8, 64.7, 64.5), whereas SimRAG, whose selection does not depend on the student, drifts down in medicine when iterated (63.7$\rightarrow$63.0) and gains only 1.0 point on CS (61.8$\rightarrow$62.8), staying 5.5 points below \method{}. With the 8B backbone the second CS round regresses on the assertion subset.

\begin{table*}[t]
\centering\scriptsize
\setlength{\tabcolsep}{4pt}
\resizebox{\linewidth}{!}{\begin{tabular}{lccccccccc}
\toprule
\textbf{Method} & \multicolumn{7}{c}{\textit{Medical}} & \textit{CS} & \textit{Science} \\
\cmidrule(lr){2-8}\cmidrule(lr){9-9}\cmidrule(lr){10-10}
 & \textbf{PubMedQA} & \textbf{BioASQ} & \textbf{MedQA} & \textbf{MedMCQA} & \textbf{MMLU-Med} & \textbf{MedXpertQA} & \textbf{Avg.} & \textbf{CS Avg.} & \textbf{Sci. Avg.} \\
\midrule
\rowcolor{oursA} \method{} & \underline{81.60} & 91.26 & \textbf{63.08} & \underline{61.37} & \textbf{78.33} & 13.39 & \textbf{64.84} & \textbf{68.19} & \textbf{87.62} \\
\midrule
\quad w/o parametric route & \textbf{82.60} & \underline{92.07} & 60.17 & 59.22 & 75.30 & 12.90 & 63.71 & 61.91 & 84.30 \\
\quad w/o triage (= random selection) & 77.60 & \textbf{92.88} & \underline{61.51} & \textbf{61.92} & \underline{77.13} & \textbf{13.84} & \underline{64.15} & \underline{67.58} & \underline{87.51} \\
\quad w/o triage, w/ round-trip filter (= SimRAG) & \textbf{82.60} & 91.26 & 59.86 & 59.62 & 74.93 & \underline{13.76} & 63.67 & 61.83 & 85.74 \\
\quad w/ retriever tuning & 61.80 & 83.66 & 60.64 & 59.79 & 76.49 & 12.45 & 59.14 & 65.89 & 87.44 \\
\quad w/ retriever tuning, w/o parametric route & 59.60 & 83.82 & 57.42 & 55.49 & 73.55 & 11.39 & 56.88 & 62.36 & --- \\
\bottomrule
\end{tabular}
}
\caption{Ablations of \method{} (Qwen3-4B, round 1). Medical results are per dataset; for CS and science we report the average. ``w/o parametric route'' puts the gold passage in the context of the retrieval-failure examples instead of training the reader to answer despite an unhelpful context (retriever frozen); ``w/o triage'' selects uniformly from the same pool at the same budget; ``w/ retriever tuning'' fine-tunes bge-base on the retrieval-failure pairs and re-indexes.}
\label{tab:ablation}
\end{table*}

\subsection{Ablation Studies}
\label{sec:ablation}

\paragraph{Effect of the parametric route.}
The single ablation that separates \method{} from SimRAG is the treatment of the \cat{retrieval-failure} examples. In Table~\ref{tab:ablation} we keep the triage, the budget and the frozen retriever, and only replace the parametric route by the ``obvious'' alternative of inserting the gold passage into the context of these examples, so that the reader is again trained exclusively on contexts that contain the evidence. On CS-Bench this variant drops from 68.2 to 61.9, exactly the level of SimRAG (61.8) and of the Stage-I model (61.9); in the scientific domain it drops from 87.6 to 84.3, below SimRAG (85.7); and in the medical domain from 64.8 to 63.7, again the SimRAG level (63.7). Figure~\ref{fig:categories} (bottom) makes the pattern visible across all three domains: mixtures in which no training context lacks the evidence (SimRAG, the ablation) sit at or below the Stage-I model; mixtures that contain such contexts (random selection, RAFT-style with 20\% distractor-only examples, \method{} with its 20\% parametric route) sit within a point of or above the closed-book model. A reader fine-tuned only on contexts that contain the answer learns to trust the context; in a domain where the retriever misses often, that reader is worse than no retrieval at all. Round-trip filtering, introduced in SimRAG to improve label quality, has the side effect of removing precisely the examples that teach a reader when \emph{not} to trust its context.

\paragraph{Effect of the triage.}
Replacing the triage by uniform selection from the same pool (``w/o triage'') costs 0.7 points in medicine, 0.6 on CS-Bench and 0.1 in science with the 4B backbone, and 2.3 on CS-Bench with the 8B backbone; adding the round-trip filter on top (``= SimRAG'') costs a further 0.5, 5.7 and 1.8. In medicine both selections gain alike on the exam-style datasets relative to Stage-I (three-seed means within a point of each other on MedQA, MedMCQA and MMLU-Med); the difference between them is PubMedQA, which random selection lowers by 3.5 points on average (77.6, 81.0, 76.8 vs.\ 82.0 for Stage-I) and \method{} preserves (81.6, 82.6, 82.2; Figure~\ref{fig:perdataset}).

\paragraph{Effect of retriever tuning (a negative result).}
The obvious remedy for \cat{retrieval-failure}s is to fix the retriever with the failed pairs (Section~\ref{sec:retriever}). Fine-tuning bge-base on them and re-indexing each round \emph{reduced} recall@5 of source passages on the medical pool from 0.48 to 0.34 and accuracy from 64.8 to 59.1 (Table~\ref{tab:ablation}); with the round-1 reader of that run held fixed, swapping in the tuned index alone lowers PubMedQA/BioASQ/MedQA from 81.4/90.9/62.1 to 61.8/83.7/60.6 (Table~\ref{tab:retriever}). On CS the tuned retriever raised pool recall (0.87$\rightarrow$0.92) yet lowered CS-Bench accuracy (65.9 vs.\ 68.2); in science it was neutral. The failed pairs are a biased sample---many questions are legitimately not specific to their source passage---so the contrastive update teaches the encoder noise and moves it away from the benchmark queries. We therefore keep the retriever frozen; curated negatives and answer-aware positives are left as the way to make a retriever route work.

\subsection{Study on the Triage}
\label{sec:triage_study}

\begin{figure*}[t]
\centering
\includegraphics[width=0.66\textwidth]{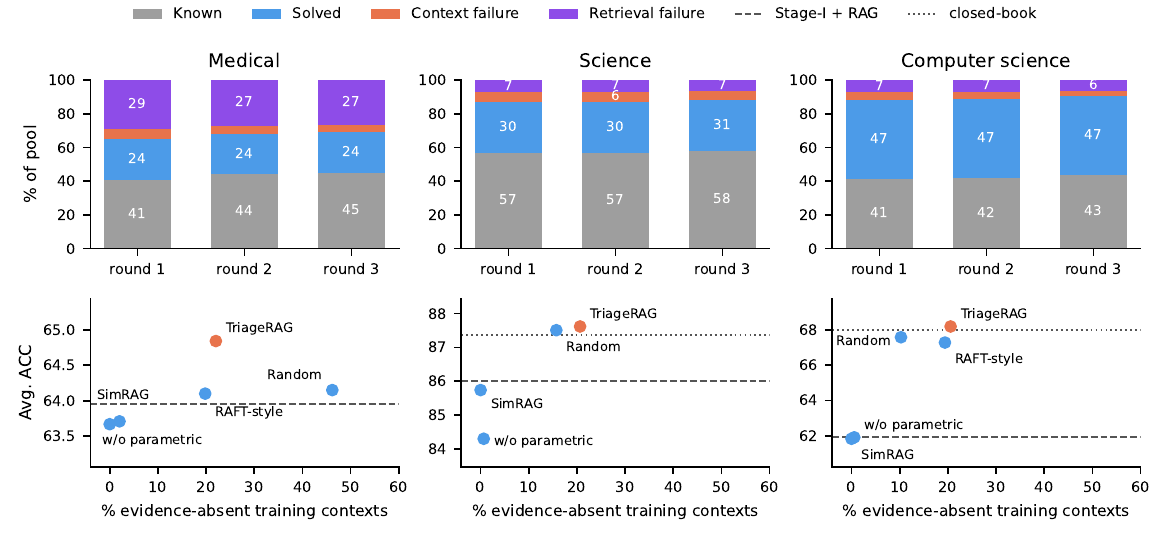}
\caption{\textbf{Top:} triage of the self-generated pool by the Qwen3-4B student across rounds (shares in \%). \textbf{Bottom:} average accuracy of each round-1 Stage-II mixture against the share of its training examples whose context does \emph{not} contain the evidence; mixtures with at most 2\% such examples (SimRAG, the ablation without the parametric route) stay at or below Stage-I, those with 10\% or more reach within a point of the closed-book level or above.}
\label{fig:categories}
\end{figure*}

\paragraph{What the triage reveals.}
Figure~\ref{fig:categories} and Table~\ref{tab:categories} (appendix) show the category shares: about 41\% of the self-generated questions in medicine and CS, and 57\% in science, are \cat{known}: the Stage-I student answers them closed-book at least three times out of four, so they carry little information, yet SimRAG spends a proportional part of its budget on them. In CS and science the retriever finds the evidence for 84--89\% of the questions and the student solves most of them; \cat{context-failure}s are 4--6\% of the pool and \cat{retrieval-failure}s 7\%. In medicine the source passage is retrieved for only 48\% of the questions and 29\% of the pool are \cat{retrieval-failure}s, so a round-trip gate discards half of the pool and, more importantly, \emph{all} the examples in which the context lacks the answer. Multiple-choice and claim items are overwhelmingly \cat{known}; 80--94\% of the failures come from short-span questions, so SimRAG's format diversity mostly supplies format practice rather than new knowledge.

\paragraph{Why round-trip filtering hurts.}
Figure~\ref{fig:categories} (bottom) plots each mixture's accuracy against the share of its examples whose training context lacks the evidence: zero for SimRAG by construction and at most 2\% for the ablation without the parametric route; the miss rate of the sampled tuples for random selection (46\% in medicine, 16\% in science, 10\% in CS); 20\% for RAFT-style and 21--22\% for \method{}. In the two domains where naive retrieval hurts, this share separates the mixtures into two groups---at most 2\% (Stage-I level) and 10\% or more (within a point of the closed-book level or above)---and the ablation falls onto the SimRAG point. The same mechanism fits the per-dataset pattern (Figure~\ref{fig:perdataset}): exam questions are the case in which the retrieved passages are topically related but lack the answer, so the reader must combine them with what it knows.

\subsection{Robustness: Training Seeds and Confidence Intervals}
\label{sec:seeds}
Every cell in Tables~\ref{tab:medical}--\ref{tab:cs} is one training run; to check that the ordering is not a seed artefact we repeated the four headline Stage-II cells---SimRAG, random selection, \method{} and \method{} without the parametric route---with two further seeds (mixture sampling and LoRA initialisation) in every domain, and the three CS-Bench cells with the 8B backbone (Table~\ref{tab:seeds}, Appendix~\ref{app:extra}). On every seed in every domain \method{} beats SimRAG and the ablation without the parametric route; \method{} vs.\ random selection and the ablation vs.\ SimRAG stay within two points. On three-seed means \method{} leads SimRAG by $+5.1$, $+1.8$ and $+2.8$ points on CS, medicine and science; removing the parametric route costs $5.8$, $1.1$ and $2.9$ points and lands within $0.8$ points of the SimRAG mean each time (seed standard deviations 0.1--1.2). Paired bootstrap intervals over test questions (10,000 resamples, seed-0 cells) agree: \method{} vs.\ SimRAG $+6.4$ \ci{4.1}{8.6} on CS, $+1.2$ \ci{0.5}{1.9} in medicine and $+1.9$ \ci{1.2}{2.5} in science, all $p<10^{-3}$; \method{} vs.\ its ablation without the parametric route $+5.8$ \ci{3.5}{8.2} on CS; \method{} vs.\ random selection significant only in medicine at 4B ($+0.7$ \ci{0.1}{1.3}) but $+2.3$ \ci{0.2}{4.4} at 8B, where the three-seed gap is $+2.0$ points ($69.9 \pm 0.4$ vs.\ $68.0 \pm 0.2$).

\section{Conclusion}
We introduce \method{}, a failure-aware self-training framework for adapting retrieval-augmented LLMs to specialized domains from an unlabeled corpus: it probes the student closed-book and open-book on every self-generated question, triages it by failure type and routes it to the training signal that fits. Across 11 datasets in three domains, and at two backbone sizes on CS-Bench, it consistently outperforms a budget-matched SimRAG, and a single ablation explains why: round-trip filtering removes every training context that lacks the evidence, and a reader that never sees such contexts over-trusts noisy retrieval. The triage doubles as a label-free diagnostic of where a domain gap lies; making retrieval failures useful to the retriever itself remains open.

\section{Limitations}
Our rounds re-probe a pool synthesized once, so the informative examples are consumed early and accuracy plateaus; regenerating the pool between rounds is future work. Our 4B/8B backbones, 16k-example Stage-I blend and PubMed slice make absolute numbers incomparable with SimRAG's 8B/27B models on full PubMed, textbooks and Wikipedia, and MedXpertQA is at chance for every configuration. Headline cells have three seeds and paired intervals; the other cells are single runs, and CS-Bench differences below one point should be read as ties. The parametric route teaches the reader to answer, not to abstain, when the context is unhelpful, and our negative retriever-tuning result covers one recipe (Appendix~\ref{app:limits} expands each point).

\subsection*{AI use statement}
Large language models were used as writing and coding assistants: to draft and edit prose, to generate analysis and plotting scripts, and to port the manuscript between templates. All experimental designs, numbers and claims were produced by the authors' code and checked by the authors against the result artifacts; LLM-generated code was tested before use. We have reviewed all AI-assisted text and code and take responsibility for the final content of this work, including any text, claims or artifacts produced with the aid of generative AI.

\subsection*{Reproducibility statement}
All datasets, corpora and backbones are public (Section~\ref{sec:setup}, Appendix~\ref{app:testdata}); prompts are in Appendix~\ref{app:prompts}, Stage-I data in Appendix~\ref{app:stage1}, and baseline and synthesis details in Appendix~\ref{app:baselines}. The code (data preparation, synthesis, probing, triage, mixture construction, LoRA training, evaluation and the paired-bootstrap script), the exact job lists that produced every cell, and per-example predictions for every reported run will be released; every number in the paper is generated from the summary files by one script.

\bibliographystyle{iclr2027_conference}
\bibliography{references}

\appendix

\section{Stage-I Data}
\label{app:stage1}
The Stage-I training set is a uniform sample of 16,000 examples from a 20,930-example general blend that mirrors the three components of SimRAG's Stage-I data: (1) context-aware QA---SQuAD \cite{rajpurkar2016squad} short-span questions (3,000), BoolQ \cite{clark2019boolq} yes/no questions (3,000) and OpenBookQA \cite{mihaylov2018openbookqa} multiple-choice questions with the supporting fact as context (3,000); (2) answer-span generation---SQuAD passages with all their annotated answer spans as the target (2,930); and (3) question generation---SQuAD answer-conditioned question generation (3,000) and Natural Questions \cite{kwiatkowski2019nq} question generation from a passage without a given answer (6,000). We omit SimRAG's general instruction-following data because our backbones are already instruction-tuned. One epoch of LoRA training (250 steps at global batch 64) took 21 minutes on four A100s for the 4B backbone and 67 minutes on two A100s for the 8B backbone. The 2,000 general replay examples used in every Stage-II round are context-QA examples drawn from the same blend.

\section{Test Data Details}
\label{app:testdata}
We evaluate on 11 datasets. (1) \textbf{Medical}:
\begin{itemize}[leftmargin=*,itemsep=0pt,topsep=1pt]
\item \textbf{PubMedQA} \cite{jin2019pubmedqa}: 500 yes/no/maybe questions about PubMed abstracts (MIRAGE split).
\item \textbf{BioASQ} \cite{tsatsaronis2015bioasq}: 618 yes/no questions from biomedical literature (MIRAGE split).
\item \textbf{MedQA} \cite{jin2021medqa}: 1,273 four-option questions from the US Medical Licensing Examination.
\item \textbf{MedMCQA} \cite{pal2022medmcqa}: the 4,183-question development set of Indian medical entrance-exam questions.
\item \textbf{MMLU-Med} \cite{hendrycks2021mmlu}: 1,089 questions from the six biomedical MMLU subjects (anatomy, clinical knowledge, professional medicine, medical genetics, college medicine, college biology).
\item \textbf{MedXpertQA} \cite{zuo2025medxpertqa}: the 2,450 text-only questions of the expert-level benchmark, with up to ten options.
\end{itemize}
(2) \textbf{Scientific}: \textbf{ARC-Challenge} \cite{clark2018arc} (1,172 test questions), \textbf{SciQ} \cite{welbl2017sciq} (1,000 test questions, kept as four-way multiple choice so that all tasks share one metric), and \textbf{MMLU-Sci} (3,013 questions from the 14 scientific MMLU subjects used by SimRAG). (3) \textbf{Computer science}: the English multiple-choice (1,336) and assertion (442) questions of \textbf{CS-Bench} \cite{song2024csbench}; we omit the fill-in-the-blank and open-ended subsets, which require an LLM judge.

\paragraph{Corpora.} Medical: MedRAG textbook snippets of at least 25 words \cite{xiong2024mirage} (123,799 of 125,847), 40 of the 1,166 MedRAG PubMed chunk files spread over the PMID range (822,980 abstracts of at least 40 words) and the abstracts of the 3,967 PMIDs referenced by PubMedQA and BioASQ retrieved from NCBI E-utilities, mirroring MedCorp in which the full PubMed is available; 950,746 passages in total. Science: SciQ training and validation supports and OpenBookQA facts (10,863 passages). Computer science: the Wikipedia computer-science subset used by SimRAG, chunked to 180 words (6,237 passages). Indices are flat inner-product FAISS indices over bge-base-en-v1.5 embeddings.

\section{Baseline and Synthesis Details}
\label{app:baselines}
\paragraph{Synthesis.} Passages with fewer than 40 words are skipped. Candidate answer spans must appear verbatim in the passage and have at most eight words. Generated questions must end with a question mark and must not contain the answer string. Multiple-choice distractors are candidate spans from other passages that do not occur in the source passage, matched on the presence of digits and within three words of the answer's length; option order is shuffled. Claims alternate between supported and refuted per passage. Self-answerability uses greedy decoding with the source passage as the only context. The pools contain 40,540 (medical; from 12,000 passages), 32,686 (science; 8,000 passages) and 20,669 (CS; 5,089 passages: 11,908 span, 4,997 multiple-choice, 3,764 claim) self-answerable tuples.
\paragraph{SimRAG$^\ddagger$.} Tuples whose source passage is among the top-5 retrieved passages for the generated question are kept (48\%, 78\% and 87\% of the medical, science and CS pools); $B$ of them are sampled uniformly and trained with the retrieved passages as context, plus the 2,000 general replay examples.
\paragraph{Random selection.} $B$ tuples are sampled uniformly from the full pool and trained with the retrieved passages as context.
\paragraph{RAFT-style$^\ddagger$.} $B$ tuples are sampled uniformly; for 80\% the gold passage replaces one of the retrieved passages at a random position, for 20\% only retrieved passages other than the gold one are used. Targets are the short answers rather than GPT-4 chain-of-thought answers, so that all methods share the same label space.
\paragraph{Realized \method{} mixture.} With the 60/20/15/5\% quotas, the round-1 mixtures contain 2,324 \cat{context-failure} (all of them), 2,400 \cat{retrieval-failure}, 6,676 \cat{solved} and 600 \cat{known} tuples in medicine ($B{=}12{,}000$; 19/20/56/5\%); 1,960 (all), 1,200, 2,540 and 300 in science (33/20/42/5\%); and 914 (all), 1,200, 3,586 and 300 on CS ($B{=}6{,}000$; 15/20/60/5\%), plus the 2,000 general replay examples in every case. The measured share of training contexts that lack the evidence is 22.0\%, 20.7\% and 20.6\% (medicine, science, CS) for \method{}, 46.2\%, 15.7\% and 10.3\% for random selection, 19.8\% and 19.4\% for RAFT-style (medicine, CS), and 2.0\%, 0.7\% and 0.6\% for the ablation without the parametric route.
\paragraph{Retriever tuning.} bge-base-en-v1.5 is fine-tuned for one epoch on the round's \cat{retrieval-failure} pairs (question, source passage) with a multiple-negatives ranking loss over in-batch negatives plus one mined hard negative per pair, the top-ranked retrieved passage other than the source (batch 64, learning rate $2\times10^{-5}$, maximum length 256); the corpus is then re-embedded and re-indexed.

\section{Additional Analyses}
\label{app:extra}
\begin{table}[h]
\centering\footnotesize
\setlength{\tabcolsep}{4pt}
\begin{tabular}{llcccc}
\toprule
\textbf{Domain / backbone} & \textbf{Method} & \textbf{seed 0} & \textbf{seed 1} & \textbf{seed 2} & \textbf{mean $\pm$ sd} \\
\midrule
\multirow{4}{*}{CS-Bench, Qwen3-4B} & SimRAG$^\ddagger$ & 61.83 & 62.25 & 63.01 & 62.36 $\pm$ 0.59 \\
 & Random selection & 67.58 & 65.47 & 67.58 & 66.88 $\pm$ 1.22 \\
 & \method{} w/o parametric route & 61.91 & 61.31 & 61.84 & 61.69 $\pm$ 0.33 \\
 & \cellcolor{oursA}\method{} & \cellcolor{oursA}68.19 & \cellcolor{oursA}67.09 & \cellcolor{oursA}67.13 & \cellcolor{oursA}\textbf{67.47} $\pm$ 0.62 \\
\midrule
\multirow{4}{*}{Medical, Qwen3-4B} & SimRAG$^\ddagger$ & 63.67 & 62.51 & 62.92 & 63.03 $\pm$ 0.59 \\
 & Random selection & 64.15 & 65.04 & 63.78 & 64.32 $\pm$ 0.65 \\
 & \method{} w/o parametric route & 63.71 & 63.58 & 64.03 & 63.77 $\pm$ 0.23 \\
 & \cellcolor{oursA}\method{} & \cellcolor{oursA}64.84 & \cellcolor{oursA}64.51 & \cellcolor{oursA}65.27 & \cellcolor{oursA}\textbf{64.87} $\pm$ 0.38 \\
\midrule
\multirow{4}{*}{Science, Qwen3-4B} & SimRAG$^\ddagger$ & 85.74 & 83.94 & 84.99 & 84.89 $\pm$ 0.90 \\
 & Random selection & 87.51 & 87.71 & 87.78 & 87.67 $\pm$ 0.14 \\
 & \method{} w/o parametric route & 84.30 & 85.75 & 84.37 & 84.81 $\pm$ 0.82 \\
 & \cellcolor{oursA}\method{} & \cellcolor{oursA}87.62 & \cellcolor{oursA}87.68 & \cellcolor{oursA}87.84 & \cellcolor{oursA}\textbf{87.71} $\pm$ 0.11 \\
\midrule
\multirow{3}{*}{CS-Bench, Qwen3-8B} & SimRAG$^\ddagger$ & 65.68 & 65.82 & 65.78 & 65.76 $\pm$ 0.07 \\
 & Random selection & 67.73 & 67.95 & 68.18 & 67.95 $\pm$ 0.23 \\
 & \cellcolor{oursA}\method{} & \cellcolor{oursA}70.06 & \cellcolor{oursA}69.43 & \cellcolor{oursA}70.22 & \cellcolor{oursA}\textbf{69.90} $\pm$ 0.42 \\
\bottomrule
\end{tabular}
\caption{Average accuracy (\%) of the round-1 Stage-II cells over three training seeds (seed 0 is the run reported in Tables~\ref{tab:medical}--\ref{tab:cs}). Seeds change the mixture sample and the LoRA initialisation; the Stage-I model, pool and round-1 probe are shared. On every seed \method{} beats SimRAG and the ablation without the parametric route; \method{} and random selection, and likewise the ablation and SimRAG, are within two points of each other.}
\label{tab:seeds}
\end{table}
\paragraph{Case studies.}

Table~\ref{tab:cases} (appendix) shows one self-generated medical question per category: a textbook fact the student recites without context (\cat{known}); a survival statistic answered only once the passage is present (\cat{solved}); a question whose evidence is in the top-5 passages while the top-ranked passage supports a competing answer (\cat{context-failure}); and a rare complication whose abstract is not retrieved, leaving topically related but unhelpful passages (\cat{retrieval-failure})---the one example of the four that SimRAG's filter discards.

\begin{figure*}[h]
\centering
\includegraphics[width=\textwidth]{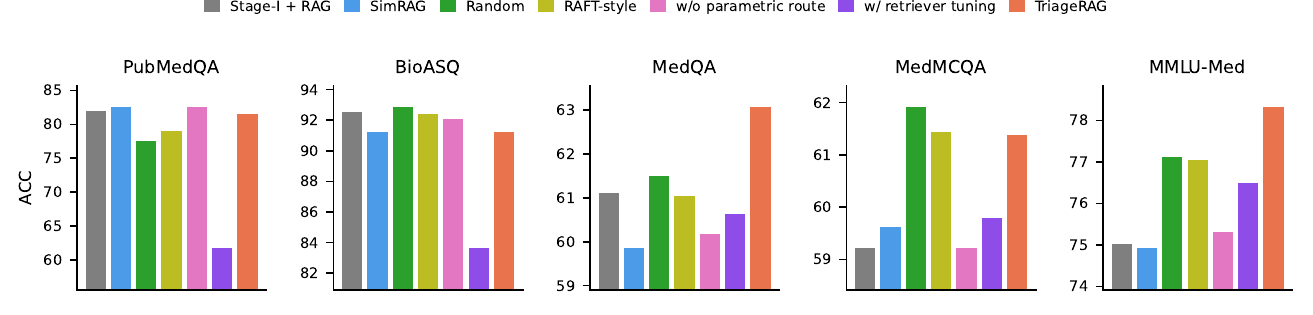}
\caption{Per-dataset accuracy of the Stage-II variants in the medical domain (Qwen3-4B, round 1). Note the different $y$-ranges. The literature yes/no datasets (PubMedQA, BioASQ) move by at most 1.3 points under SimRAG, the ablation without the parametric route and \method{}; random selection ($-4.4$) and RAFT-style ($-3.0$) lower PubMedQA, and retriever tuning damages both. The exam-style datasets (MedQA, MedMCQA, MMLU-Med) carry the gains of \method{} over Stage-I.}
\label{fig:perdataset}
\end{figure*}

Table~\ref{tab:categories} gives the round-1 triage statistics with a per-format breakdown, Table~\ref{tab:retriever} the controlled index comparison behind the retriever-tuning result, and Figure~\ref{fig:perdataset} the per-dataset view of the medical ablations.

\begin{table}[h]
\centering\scriptsize
\setlength{\tabcolsep}{3pt}
\resizebox{\textwidth}{!}{\begin{tabular}{lrrrrrrr}
\toprule
\textbf{Pool} & \textbf{Size} & \textbf{Known} & \textbf{Solved} & \textbf{Ctx.F.} & \textbf{Ret.F.} & \textbf{Hit@5} & \textbf{R@5} \\
\midrule
Medical & 40,540 & 40.9 & 24.2 & 5.7 & 29.2 & 53.4 & 48.1 \\
\quad span & 19,553 & 4.5 & 36.1 & 11.0 & 48.4 & & \\
\quad MC & 11,775 & 75.9 & 8.8 & 0.1 & 15.2 & & \\
\quad claim & 9,212 & 73.3 & 18.6 & 1.8 & 6.3 & & \\
Science & 32,686 & 56.6 & 30.3 & 6.0 & 7.0 & 84.2 & 77.9 \\
\quad span & 18,622 & 31.5 & 47.7 & 9.9 & 11.0 & & \\
\quad MC & 7,845 & 93.2 & 4.9 & 0.2 & 1.7 & & \\
\quad claim & 6,219 & 85.8 & 10.5 & 1.8 & 1.9 & & \\
CS & 20,669 & 41.2 & 47.2 & 4.4 & 7.1 & 88.6 & 87.0 \\
\quad span & 11,908 & 12.4 & 70.3 & 6.6 & 10.6 & & \\
\quad MC & 4,997 & 84.8 & 12.3 & 0.2 & 2.7 & & \\
\quad claim & 3,764 & 74.7 & 20.5 & 3.2 & 1.7 & & \\
\bottomrule
\end{tabular}
}
\caption{Round-1 triage of the self-generated pools (Qwen3-4B Stage-I student, frozen retriever; shares in \%). Hit@5: fraction of questions whose evidence is in the top-5 passages ($h{=}1$); R@5: fraction whose source passage is among them.}
\label{tab:categories}
\end{table}

\begin{table}[h]
\centering\small
\setlength{\tabcolsep}{4pt}
\begin{tabular}{lccc}
\toprule
\textbf{Index at test time} & \textbf{PubMedQA} & \textbf{BioASQ} & \textbf{MedQA} \\
\midrule
Base bge-base & \textbf{81.4} & \textbf{90.9} & \textbf{62.1} \\
Tuned on round-1 failures & 61.8 & 83.7 & 60.6 \\
\midrule
Recall@5 on the pool & \multicolumn{3}{c}{0.48 (base) $\rightarrow$ 0.34 (tuned)} \\
\bottomrule
\end{tabular}
\caption{Controlled comparison holding one round-1 medical reader fixed (Qwen3-4B; the round-1 reader of the retriever-tuning run, trained on the same round-1 mixture as \method{} in a separate run, hence the small differences from Table~\ref{tab:medical}): fine-tuning the retriever on retrieval-failure pairs degrades both recall of the source passage and downstream accuracy.}
\label{tab:retriever}
\end{table}

\begin{table*}[h]
\centering\scriptsize
\setlength{\tabcolsep}{4pt}
\renewcommand{\arraystretch}{1.15}
\begin{tabular}{p{2.3cm}p{11.05cm}}
\toprule
\multicolumn{2}{p{13.63cm}}{\centering \textbf{\cat{known}} \quad ($\pcb{=}1.00$, $\pob{=}1.00$, source passage retrieved at rank 1) \quad $\rightarrow$ 5\% consolidation share}\tabularnewline
\midrule
\textbf{Question / Answer} & Where does the lymphatic system filter the lymph? \quad \textbf{A:} lymph nodes \\
\textbf{Source (Textbooks)} & \textit{Physiology, Levy.} In addition to returning fluid and protein to the vascular bed, the lymphatic system filters the lymph at the \evid{lymph nodes} and removes foreign particles such as bacteria. \\
\textbf{Top retrieved} & (same passage) \\
\midrule
\multicolumn{2}{p{13.63cm}}{\centering \textbf{\cat{solved}} \quad ($\pcb{=}0.00$, $\pob{=}1.00$, source passage retrieved at rank 1) \quad $\rightarrow$ 15\% consolidation share}\tabularnewline
\midrule
\textbf{Question / Answer} & What was the 10-year survival rate for relapsing polychondritis? \quad \textbf{A:} 55\% \\
\textbf{Source (Textbooks)} & \textit{Internal Medicine, Harrison.} The course of relapsing polychondritis is highly variable. \ldots\ In one study, the 5-year estimated survival rate was 74\% and the \evid{10-year survival rate was 55\%}. \\
\textbf{Top retrieved} & (same passage) \\
\midrule
\multicolumn{2}{p{13.63cm}}{\centering \textbf{\cat{context-failure}} \quad ($\pcb{=}0.50$, $\pob{=}0.00$, evidence in the top-5 passages, source passage not retrieved) \quad $\rightarrow$ contextual SFT, weight $\propto 1-\pob$}\tabularnewline
\midrule
\textbf{Question / Answer} & What type of cancer do long chain n-3 PUFA help to prevent? \quad \textbf{A:} breast cancer \\
\textbf{Source (PubMed)} & \textit{Endogenous synthesis of n-3 polyunsaturated fatty acids in Fat-1 mice \ldots} Long chain n-3 PUFA have been shown to have chemopreventive properties against \evid{breast cancer} through various mechanisms. \\
\textbf{Top retrieved} & \textit{Dietary n-3/long-chain n-3 polyunsaturated fatty acids for prevention of sporadic \miss{colorectal} tumors: a randomized controlled trial \ldots} To address preventive effects of n-3 PUFAs on CRTs, a randomized controlled trial was conducted. \quad (the answer string appears only in a lower-ranked passage; the top passage supports a competing answer) \\
\midrule
\multicolumn{2}{p{13.63cm}}{\centering \textbf{\cat{retrieval-failure}} \quad ($\pcb{=}0.00$, $\pob{=}0.00$, evidence absent from the top-5 passages) \quad $\rightarrow$ parametric SFT with the retrieved (unhelpful) context}\tabularnewline
\midrule
\textbf{Question / Answer} & What is a rare complication after renal transplantation? \quad \textbf{A:} iliac artery stenosis \\
\textbf{Source (PubMed)} & \textit{Iliac artery stenosis as a cause of posttransplant renal failure and claudication.} \miss{Iliac artery stenosis (IAS) is a rare complication after renal transplantation.} We demonstrate a case of ipsilateral external IAS proximally to the anastomosis in a kidney recipient \ldots \\
\textbf{Top retrieved} & \textit{Ultrasound evaluation of uncommon renal transplant complications.} Twenty-five patients who had undergone renal transplants developed uncommon complications that were detected and followed serially with sonography. Two patients developed renal calculi \ldots \quad (topically related, answer absent) \\
\bottomrule
\end{tabular}
\caption{Case studies: one self-generated medical question per triage category with its probe outcomes ($\pcb$, $\pob$, evidence hit) and the route it receives. \evid{Green} marks evidence present in the retrieved passages; \miss{red} marks the answer-bearing sentence of the source passage when it was \emph{not} retrieved.}
\label{tab:cases}
\end{table*}

\section{Additional Results}
\paragraph{Effect of rounds.}
Figure~\ref{fig:rounds} in Appendix~\ref{app:rounds} plots the average accuracy per round (all numbers in Table~\ref{tab:rounds}). Because \method{} re-probes the pool with the updated student, the mixture shifts: in the medical domain the \cat{known} share rises from 40.9\% to 44.2\% and 45.2\% over the three rounds and the \cat{retrieval-failure} share falls from 29.2\% to 27.1\% and 26.8\%, i.e.\ the student has absorbed part of what it was trained on; on CS-Bench the \cat{context-failure} share falls from 4.4\% to 2.7\%. Accuracy, however, moves little after the first round: medicine plateaus (64.8, 64.7, 64.5), science peaks at round two (87.6, 88.0, 87.5) and so does CS-Bench (68.2, 68.8, 68.3). With a fixed pool the informative examples are largely consumed in round one; the retriever-tuned variant is the exception on CS (65.9, 67.7, 68.0), where later rounds partially recover from the damaged index. Iterating SimRAG lowers medical accuracy (63.7, 63.5, 63.0) and gains at most 1.5 points on CS (61.8, 63.4, 62.8), never approaching \method{}. Re-probing is therefore most useful when the pool is refreshed between rounds.

\label{app:rounds}
Table~\ref{tab:rounds} lists the average accuracy of every iterated method per round for all domains and backbones, and Figure~\ref{fig:rounds} plots them for the 4B backbone.

\begin{figure*}[h]
\centering
\includegraphics[width=\textwidth]{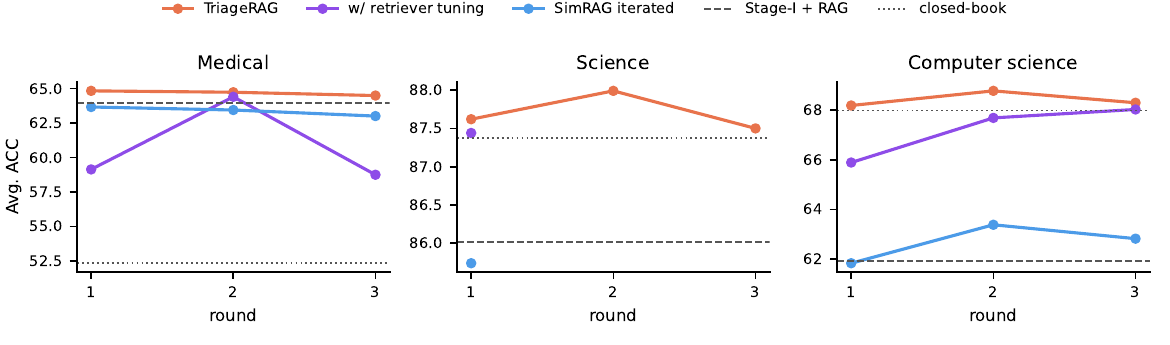}
\caption{Average accuracy per round (Qwen3-4B). Re-probing the updated student changes the mixture of \method{} every round; SimRAG's selection does not depend on the student, so its later rounds re-train on the same distribution.}
\label{fig:rounds}
\end{figure*}

\begin{table*}[h]
\centering\small
\setlength{\tabcolsep}{6pt}
\resizebox{\textwidth}{!}{\begin{tabular}{llccc}
\toprule
\textbf{Domain / backbone} & \textbf{Method} & \textbf{Round 1} & \textbf{Round 2} & \textbf{Round 3} \\
\midrule
Medical / Qwen3-4B-Instruct & \method{} & 64.84 & 64.74 & 64.50 \\
 & \method{} w/ retriever tuning & 59.14 & 64.41 & 58.75 \\
 & SimRAG$^\ddagger$ iterated & 63.67 & 63.45 & 63.01 \\
\midrule
Science / Qwen3-4B-Instruct & \method{} & 87.62 & 87.99 & 87.50 \\
 & \method{} w/ retriever tuning & 87.44 & --- & --- \\
 & SimRAG$^\ddagger$ iterated & 85.74 & --- & --- \\
\midrule
Computer science / Qwen3-4B-Instruct & \method{} & 68.19 & 68.78 & 68.30 \\
 & \method{} w/ retriever tuning & 65.89 & 67.69 & 68.03 \\
 & SimRAG$^\ddagger$ iterated & 61.83 & 63.38 & 62.82 \\
\midrule
Computer science / Qwen3-8B & \method{} & 70.06 & 67.73 & --- \\
 & SimRAG$^\ddagger$ iterated & 65.68 & --- & --- \\
\bottomrule
\end{tabular}
}
\caption{Average accuracy per round for the iterated methods.}
\label{tab:rounds}
\end{table*}

\paragraph{Answer parsing.}
\label{app:parse}
Letters are parsed from the first line of the output (patterns \texttt{A}, \texttt{(A)}, \texttt{A.}, ``answer is A''); yes/no/maybe and true/false are matched as whole words. Unparsable outputs are counted as wrong. Across all evaluations of fine-tuned models the number of unparsable outputs was at most 3 per dataset; the untuned backbone with retrieval produced 86 unparsable outputs on MedMCQA and 32 on CS-Bench MC (it sometimes answers with a sentence instead of a letter), which slightly disadvantages the ``+ RAG'' rows.

\section{Extended Limitations}
\label{app:limits}
\paragraph{Fixed pool and saturation.} Our rounds re-probe a pool that is synthesized once; the informative examples are largely consumed in the first round and accuracy plateaus. Regenerating the pool from new passages between rounds, as the shifting triage suggests, is left to future work.
\paragraph{Scale and corpora.} We use 4B and 8B backbones, a Stage-I blend of 16k examples and a PubMed slice rather than all of PubMed, so absolute numbers are not comparable with SimRAG's 8B/27B models, which use full PubMed, textbooks and Wikipedia; all our comparisons are internal and budget-matched. MedXpertQA is at chance level for every configuration we trained and cannot separate methods at this scale.
\paragraph{Seeds and intervals.} The headline round-1 cells were replicated with three training seeds and every comparison in the main text carries a paired bootstrap interval; the remaining cells (rounds 2--3, RAFT-style and the retriever-tuned variants) are single runs, and CS-Bench differences of one point or less between the strong variants should be read as ties.
\paragraph{Abstention and calibration.} The parametric route teaches the reader to produce a verified answer when the context is unhelpful; it does not teach abstention, and we did not measure calibration or behaviour on unanswerable questions.
\paragraph{Retriever route.} Our negative result covers one reasonable recipe (contrastive fine-tuning of bge-base on failed pairs with retrieved hard negatives). Curated negatives, smaller learning rates or mixing with the original training data may reverse it.

\section{Prompt Details}
\label{app:prompts}
\paragraph{Inference and training.} System message: ``You are a helpful, knowledgeable assistant. Use the provided context when it is relevant.'' User message:
\begin{quote}\small\ttfamily
Context:\\{[1] passage 1 \ldots [5] passage 5}\\[3pt]
\{dataset-specific instruction, e.g.\ ``Answer the following question by selecting one of the provided options with A, B, C, or D. Please answer with the capitalized alphabet only, without adding any extra phrase or period.''\}\\[3pt]
Question: \{question\}\\A. \ldots\ B. \ldots\ C. \ldots\ D. \ldots
\end{quote}
Closed-book probing uses the same message without the context block.
\paragraph{Answer generation.}
\begin{quote}\small\ttfamily
Context:\\\{passage\}\\[3pt]Based on the context, generate several candidate spans within the passage that are likely to be answers to a question. The answers can be entities, verbs or even numbers. Make sure that the answers are different and diverse. Separate different candidate answers with a semicolon (`;'). Output only the candidate answers.
\end{quote}
\paragraph{Question generation.}
\begin{quote}\small\ttfamily
Context:\\\{passage\}\\[3pt]Based on the context, please generate a question that is relevant to the information provided. The question should stand alone and not refer back to the context explicitly. The question should be clear and understandable without needing the context. The answer to the question should be: \{answer\}\\Output only the question.
\end{quote}
\paragraph{Claim generation.}
\begin{quote}\small\ttfamily
Context:\\\{passage\}\\[3pt]Based on the context, please write one specific factual claim that is \{supported | refuted\} by the context. The claim should stand alone and be understandable without the context. Output only the claim.
\end{quote}

\end{document}